\documentclass[10pt,twocolumn,letterpaper]{article}

\usepackage[pagenumbers]{arxiv} 

\definecolor{arxivblue}{rgb}{0.21,0.49,0.74}
\usepackage[pagebackref,breaklinks,colorlinks,allcolors=arxivblue]{hyperref}

\usepackage{arydshln}
\usepackage{color}
\usepackage{colortbl}
\usepackage{graphicx}
\usepackage{algorithm}
\usepackage{algorithmic}

\usepackage{multirow}
\usepackage{amssymb}
\usepackage{booktabs}
\usepackage{adjustbox}
\usepackage{multirow}
\usepackage{pifont}
\usepackage{bbding}
\usepackage{makecell}

\usepackage{CJK}

\makeatletter
\newcommand{\thickhline}{
    \noalign {\ifnum 0=`}\fi \hrule height 1pt
    \futurelet \reserved@a \@xhline
}
\makeatother

\newcommand*\samethanks[1][\value{footnote}]{\footnotemark[#1]}

\title{From Image to Video, what do we need in multimodal LLMs?}

\author{
  Suyuan Huang$^{1}$\thanks{Equal contribution},\quad Haoxin Zhang$^{2}$\samethanks,\quad Linqing Zhong$^{3}$\samethanks,\quad Honggu Chen$^{1}$,\\
  Yan Gao$^{2}$,\quad Yao Hu$^{2}$,\quad Zengchang Qin$^{1,4}$\thanks{Corresponding author} \\[6pt]
  {
      $^{1}$Intelligent Computing and Machine Learning Lab, School of ASEE, Beihang University
  }\\
  {
      $^{2}$Xiaohongshu\quad
      $^{3}$School of Sino-French Engineer, Beihang University
  }\\
  {
      $^{4}$College of Engineering and Computer Science, VinUniversity
  }
}

\begin{document}
\maketitle
\begin{abstract}
Covering from Image LLMs to the more complex Video LLMs, the Multimodal Large Language Models (MLLMs) have demonstrated profound capabilities in comprehending cross-modal information as numerous studies have illustrated. Previous methods delve into designing comprehensive Video LLMs through integrating video foundation models with primitive LLMs. Despite its effectiveness, such paradigm renders Video LLM's structure verbose and typically requires substantial video data for pre-training. Crucially, it neglects leveraging the foundational contributions of ready-made Image LLMs. In this paper, we introduce \textbf{RED-VILLM}, a \textbf{R}esource-\textbf{E}fficient \textbf{D}evelopment pipeline which builds robust \textbf{Vi}deo \textbf{LLM}s through leveraging the prior knowledge of Image LLMs. Specifically, since a video is naturally a combination of images along the temporal dimension, we devise a temporal adaptation plug-and-play structure, endowing the backbone Image LLM with the capability to grasp temporal information. Moreover, through applying this pipeline, we achieve the first Video LLM within the Chinese-speaking community. Extensive experiments demonstrate that Video LLMs developed through our approach surpass conventional Video LLMs, requiring minimal instructional data and training resources. Our approach highlights the potential for a more cost-effective and scalable advancement in multimodal models.
\end{abstract}
\section{Introduction}

\begin{figure}[tbp]
\centering
\includegraphics[width=\linewidth]{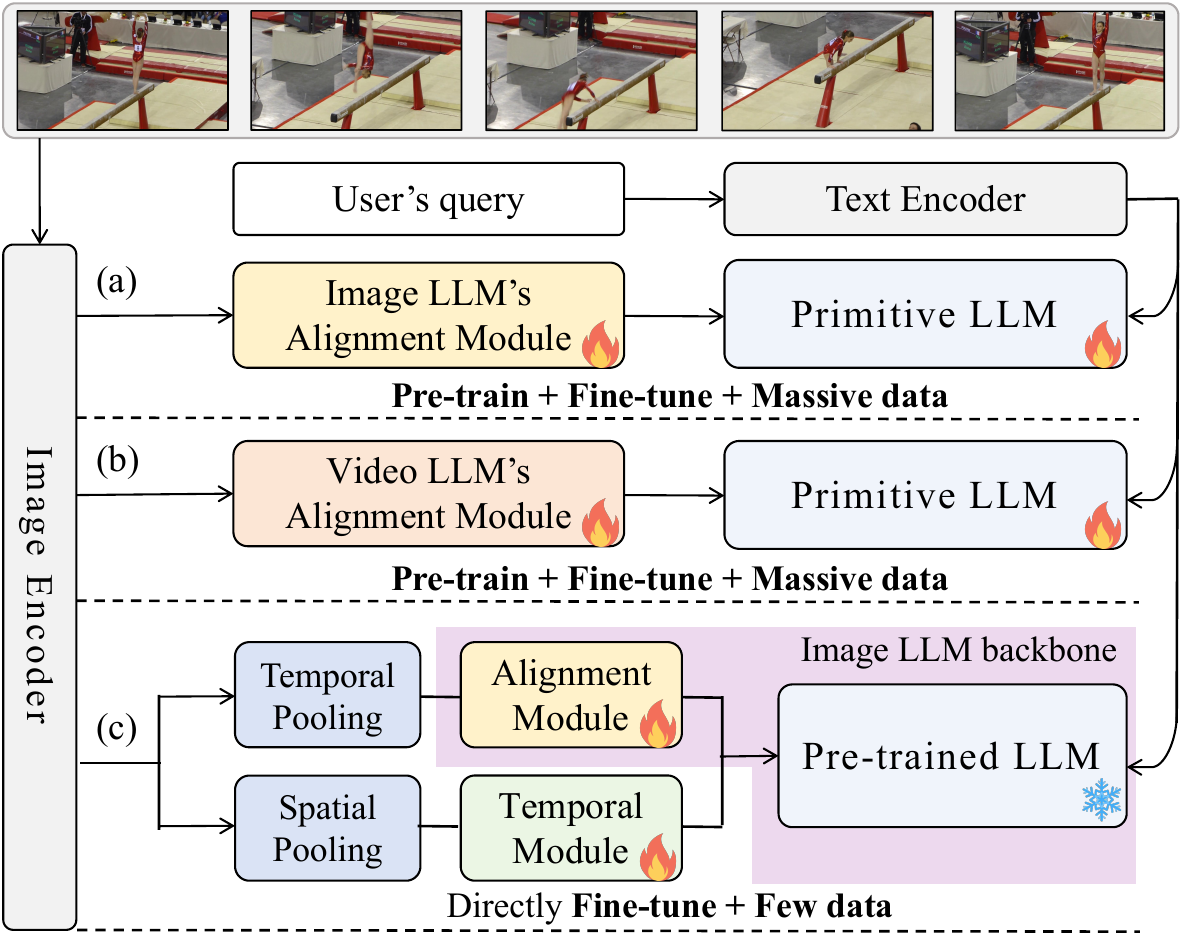}
\caption{Comparing different MLLMs' training paradigms. (a) presents the training of Image LLM. (b) shows the typical training process of Video LLM (i.e., following ``pre-train + fine-tune'' paradigm and utilizing diverse video-text data). As shown in (c), our proposed pipeline leverages the prior knowledge embedded in Image LLM backbone's pre-trained LLM to efficiently establish a robust Video LLM. Besides, the volume of data required is minimal.}
\label{fig:fig1}
\end{figure}

In recent years, with the rapid development of LLMs, transferring their outstanding generation and generalization capabilities from text-only contexts to multimodal ones has attracted a surge of interest, leading to the development of MLLMs.
The MLLMs are capable of understanding multimodal information and completing downstream tasks (e.g., visual question answering). According to the modalities that they support, the mainstream MLLMs can be divided into two categories: Image LLMs and Video LLMs. 

Typically, the MLLM comprises three components: the encoders to extract visual and textual representations, a primitive LLM for inference and crucially, the alignment module. Compared with the original LLM, the alignment module is additionally proposed because there is a significant modal gap between visual and textual information. The alignment module aligns visual features with the textual representations to bridge the gap, successfully enhancing MLLM's reasoning capacities. 

Previous methods apply various structures as alignment modules and achieve exceptional performance. Despite the alignment module's effectiveness, it often requires large-scale image-text or video-text paired data and extensive training efforts to obtain a well-performing MLLM.

As shown in Fig.~\ref{fig:fig1}(a), the training paradigm of Image LLMs contains two stages. In the first stage, large-scale datasets are utilized to render the model adapt to image features and textual features. In the process, the parameters of the alignment module and the LLM are usually adjusted for optimization. Substantially, in the second stage, a limited amount of GPT-annotated or human-annotated data are used for instruction-tuning. In this way, we realize a Image LLM which comprehends both image and text information.

Following the success of Image LLMs, numerous works focus on transferring LLMs' remarkable reasoning abilities to video domain. Most of them adapt the same two-stage training paradigm as the Image LLMs (i.e., conducting end-to-end training of video and textual modalities directly from scratch), as shown in Fig.~\ref{fig:fig1}(b). Although these efforts have  demonstrated outstanding performance, such approaches take a roundabout way. They neglect leveraging the foundational contributions of ready-made Image LLMs. 

Intuitively, since a video is an expansion of a series of images over time, a video and images both belong to visual modalities. The modal gap between them is much smaller compared to the gap between images and texts.
Besides, Image LLMs have already achieved well-aligned integration of images and texts, providing a solid foundation for aligning modalities in Video LLMs.

Therefore, a key question naturally arises: \textbf{``Can we efficiently develop Video LLMs based on Image LLMs by fully utilizing the modal alignment knowledge contained in them?''} Considering that a video serves as a temporal extension of images, we posit that this problem can be divided into two key aspects: how to extract videos' representations adapted to Image LLMs and how to extend Image LLMs' understanding of temporal information.

To address these key issues, we propose RED-VILLM, a Resource-Efficient Development pipeline which builds Video LLMs directly from Image LLMs, as shown in Fig.~\ref{fig:fig1}(c).

Specifically, since a video can be viewed as a temporal series of dense images, we propose spatial and temporal pooling methods, capturing correspondingly video's temporal and spatial features. While the spatial representations can be directly processed by pre-trained Image LLMs, we design a plug-and-play temporal adaptation module to endow the backbone Image LLM with the capability to grasp temporal information. Moreover, since our proposed pipeline performs one-stage instruction fine-tuning, it requires minimal instructional training data to build a weill-performing Video LLM. Furthermore, we implement our approach utilizing three mainstream Image LLMs (i.e., LLaVA~\cite{liu2024visual}, Qwen-VL~\cite{bai2023qwenvl} and Deepseek-VL~\cite{lu2024deepseek}). Relying on a small amount of Chinese instructional data, we develop the first Video LLM in the Chinese-speaking community.

Our contributions can be summarized as follows:
\begin{itemize}
  \item We propose a general pipeline for developing Video LLMs directly from Image LLMs. Building upon the pre-trained Image LLMs, our pipeline realizes an efficient transition with minimal training data.
  \item We introduce a plug-and-play temporal enhancement module that endows Image LLMs with the capability to understand temporal information in videos.
  \item We present the first Video LLM designed for understanding videos within the Chinese-speaking community. Compared with existing approaches, our model achieves state-of-the-art performance.
\end{itemize}
\section{Related Work}
\subsection{Large Language Models}

Large Language Models~\cite{touvron2023llama} have showcased remarkable capabilities in the field of Natural Language Processing (NLP). GPT~\cite{brown2020language} leverages generative pre-trained transformers and achieves outstanding generation capabilities. Following this paradigm, a multitude of studies~\cite{zhang2022opt, chiang2023vicuna} further propel model's emergent and generalization capabilities by scaling up data size and model size. For instance, trained on a large scale of Chinese instructions, Qwen~\cite{bai2023qwen} exhibits exceptional capabilities within the Chinese-speaking community. 

\subsection{Multimodal Large Language Models}
Building upon the significant achievements of LLMs, an increasing amount of research focus on extending the remarkable capabilities of LLMs to other modalities, enabling multimodal reasoning and action. Such models are therefore called Multimodal Large Language Models (MLLMs). The mainstream MLLMs can be categorized into two types: Image LLMs and Video LLMs.

\textbf{Image LLMs.} To enable LLMs to understand image information, the most crucial point is aligning visual and textual information. Numerous works delve this issue. Flamingo~\cite{alayrac2022flamingo} and PaLM-E~\cite{driess2023palm} have taken the lead in aligning images' visual information with LLMs' semantic space, showing outstanding performance in many downstream tasks. BLIP-2~\cite{li2023blip} introduces the Q-Former structure to extract the visual information most relevant to text. LLaVA employs a simple projection layer with few learnable parameters to achieve modal alignment. MiniGPT-4~\cite{zhu2023minigpt} and InstructBLIP~\cite{dai2024instructblip} construct high-quality instruction pairs to enhance models' performance. These models fulfill great image-text alignment and potentially serve as a foundation for expansion to other modalities.

\begin{figure*}[htbp]
\centering
\includegraphics[width=1.0\textwidth]{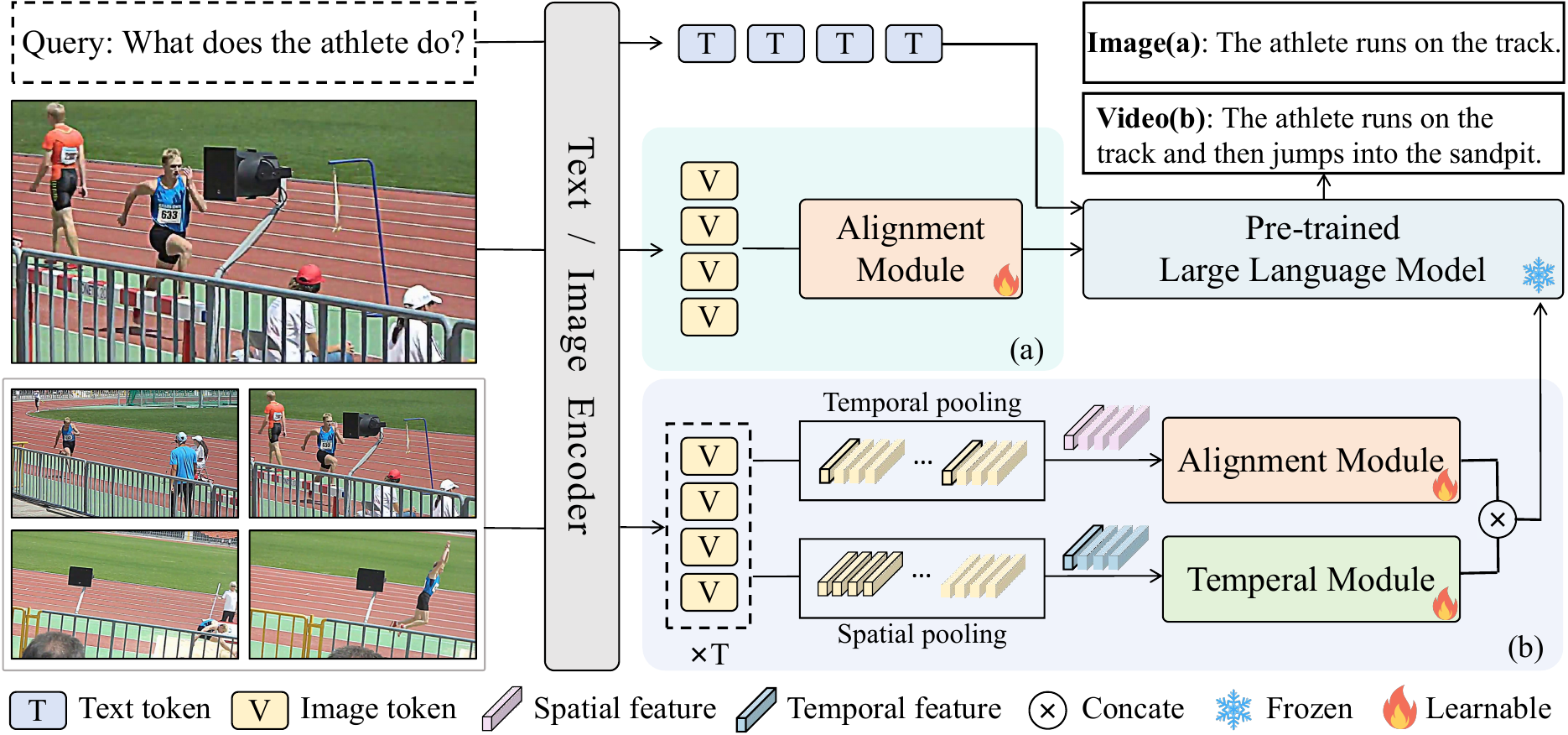}
\caption{Overview of RED-VILLM. 
(a) The Image LLM Pipeline first aligns the image features extracted by the visual encoder. Then, the aligned image tokens and text tokens are combined as input for the LLM.
(b) Our RED-VILLM preserves the main architecture of the Image LLM, adding only a plug-and-play temporal module to boost LLM’s temporal understanding. Initially, the Image LLM's visual encoder converts input frames into features, which are then pooled temporally and spatially to derive respective features. These features are concatenated after being aligned through the Image LLM’s alignment module and the temporal module. Finally, the LLM processes the combined video representation tokens and user instructions to generate responses.}
\label{fig:framework}
\end{figure*}

\textbf{Video LLMs.} Similar to Image LLMs, Video LLMs aim to support video understanding in LLMs. Following the success of Image LLMs, recent efforts~\cite{zhang2023video,li2023videochat, maaz2023video} consume diverse multimodal data and apply the two-stage training paradigm to design robust Video LLMs. These works have successfully improved LLMs' video understanding capabilities.

Despite their effectiveness, these works overlook the foundational contributions of Image LLM during training the Video LLM. Essentially, visual and textual information are already well-aligned during the pre-training of Image LLMs. We can efficiently develop a Video LLM by leveraging the prior alignment knowledge embedded in the pre-trained backbone LLM of Image LLM. In this way, both the training process and the data required to train a well-performing Video LLM are greatly reduced.

Hence, in this paper we proposee RED-VILLM, a resource-efficient pipeline to build Video LLM based on Image LLM. 
\section{Methodology}

We introduce RED-VILLM, an efficient pipeline for developing Video LLMs from Image LLMs with minimal instruction data and training parameters. The approach leverages the pre-trained alignment capabilities of Image LLMs to enable video understanding. 

\subsection{Overview}
\label{sec:overview}
Our proposed RED-VILLM is illustrated in Fig.~\ref{fig:framework}. The Image LLM aligns image and text modalities via an alignment module, then inputs the aligned tokens into the LLM. We build upon this framework to extend the LLM's capabilities to video understanding. Since video is a sequence of frames, we address two main challenges:

\begin{itemize}
\item Representing video information coherently, as videos are continuous streams.
\item Enabling the Image LLM to understand temporal information.
\end{itemize} 

Our pipeline addresses these challenges by first processing video frames with an image encoder to extract feature tokens. 
We then apply spatial and temporal pooling to obtain temporal and spatial representations, as described in Sec.~\ref{sec:feature_extraction}. 
The spatial features are aligned using the Image LLM's alignment module, while the temporal features are aligned using a plug-and-play temporal module. These spatial and temporal features are concatenated, resulting in a spatio-temporal representation that aligns with the LLM's textual semantic space, as detailed in Sec.~\ref{sec:temporal module}. Finally, the aligned video and text tokens are used for instruction-tuning, as described in Sec.~\ref{sec:instrcution tuning}.

\subsection{Video Feature Extraction}
\label{sec:feature_extraction}

We use the pre-trained image feature encoder from the base Image LLM for video feature extraction. 
Since a video can be viewed as a sequence of frames, the visual encoder, trained for image feature extraction, can naturally process video information. 
However, this encoder operates independently on each frame, without capturing temporal relationships between them. 
To address this, we modify the image features through spatial and temporal pooling to extract spatio-temporal representations.

Given a raw video, the visual input \(V_i \in \mathbb{R}^{T \times H \times W \times C}\) consists of \(T\) frames of size \(H \times W\). The image encoder processes these frames independently, producing frame-level embeddings \(X_i \in \mathbb{R}^{T \times N \times D}\), where \(N\) is the number of image patches and \(D\) is the embedding dimension.

To obtain comprehensive video representations, we first apply temporal pooling by averaging embeddings along the temporal dimension, resulting in video-level spatial representations \(z_i \in \mathbb{R}^{N \times D}\). Temporal pooling integrates information across frames, capturing the video's temporal dynamics. Next, we apply spatial pooling to the frame-level embeddings, averaging across the spatial dimension to obtain video-level temporal representations \(t_i \in \mathbb{R}^{T \times D}\), which captures the spatial interactions within frames.

This approach efficiently generates video-level spatio-temporal representations, incorporating both frame content and temporal information.

\subsection{Temporal Module}
\label{sec:temporal module}
The extracted video features need to be projected into the LLM's embedding space before being fed to it. 
The pre-trained Image LLM alignment module effectively handles spatial relationships within images but does not capture temporal dynamics across video frames.

To model these temporal relationships, we propose a Temporal Module that integrates seamlessly with the Image LLM. 

Specifically, we decouple spatial features \(z_i\) and temporal features \(t_i\). 
The Image LLM's alignment module is used to process spatial features \(z_i\). Besides, the Temporal Module aligns temporal features \(t_i\) with semantic space, producing corresponding language embedding tokens. The conversion is as follows:
\begin{equation}
Q_z = g_z(z_i) \in \mathbb{R}^{N \times K}, \quad Q_t = g_t(t_i) \in \mathbb{R}^{T \times K}.
\end{equation}

Here, \(g_z\) represents the Image LLM’s alignment module while \(g_t\) denotes the Temporal Module. \(K\) is the output dimension. The spatial and temporal features are then concatenated to form the video-level aligned tokens \(Q_v\) according to the following formula:
\begin{equation}
Q_{v} = [Q_{t}, Q_{z}] \in \mathbb{R}^{(T+N) \times K}.
\end{equation}

We find that fine-tuning the Image LLM alignment module on temporal information yields competitive performance, as the pre-trained model already captures substantial alignment. 
Thus, we integrate the Temporal Module with the base parameters of the Image LLM alignment module and fine-tune only the parameters of the temporal module, ensuring a smooth enhancement of model capabilities.

\subsection{Video Instruction Tuning}
\label{sec:instrcution tuning}
Our RED-VILLM is trained based on the Image LLM framework, which effectively utilizes the pre-trained alignment of image and text modalities. 
This pre-training allows RED-VILLM to skip modality alignment training, making it more efficient.
We employ instruction-tuning on the prediction tokens of the LLM, leveraging its autoregressive training objectives. 
During the fine-tuning stage, the instruction data are organized in the following prompt template:
\begin{equation}
    User: <Video \: Tokens><Instruction> \: Assistant:
\end{equation}

Here, $<Instruction>$ represents a randomly selected question about the video, and the prediction Answer corresponds to the specific question asked.
 The video tokens \( Q_v \) and instruction tokens \( Q_i \) are concatenated to form the LLM’s input:
\begin{equation}
    User: <Q_v><Q_i> \: Assistant:
\end{equation}

Throughout training, the weights for both the video encoder and the Image LLM remain fixed, with the model focusing on maximizing the likelihood of predicting tokens that represent the answer.
\section{Experiments}

\subsection{Experimental Setup}
\textbf{Implementation Details.}
We employ LLaVA as English Image LLM, Qwen-VL and DeepSeek-VL as Chinese Image LLMs.
We only update the alignment module, which projects the video features to the LLMs’ input space, and the plug-and-play temporal module, while the rest of the architecture is kept frozen. 
We fine-tune the model for 3 epochs using a learning rate of 2e-5 and an overall batch size of 32. During inference, for memory efficiency, we load the models in FP16 mode.

\textbf{Datasets.}
For instruction-tuning, we use 100k video-text instructions from the ActivityNet dataset, as proposed by Video-ChatGPT. 
The dataset includes various video-based question-answer pairs, covering temporal, spatial, and content-related aspects of the videos. 
For the Chinese data, we translate the English instructions into Chinese using DeepL Pro, followed by manual review and careful refinement.

\begin{table*}[h]
\centering
\caption{Comparison with leading methods on the video-based generative performance benchmark.}
\label{table: generative benchmark}
\begin{tabular}{llccccc}
\toprule
\textbf{Method} & \textbf{LLM} & \textbf{CI} & \textbf{DO} & \textbf{CU} & \textbf{TU} & \textbf{CO} \\
\midrule
Video-LLaMA~\cite{zhang2023video}      & Vicuna-7B         & 1.96 & 2.18 & 2.16 & 1.82 & 1.79 \\
LLaMA-Adapter~\cite{zhang2023llama}    & LLaMA-7B          & 2.03 & 2.32 & 2.30 & 1.98 & 2.15 \\
VideoChat~\cite{li2023videochat}       & Vicuna-7B         & 2.23 & 2.50 & 2.53 & 1.94 & 2.24 \\
Video-ChatGPT~\cite{maaz2023video}     & LLaVA-7B          & 2.40 & 2.52 & 2.62 & 1.98 & 2.37 \\
\midrule
\textbf{RED-VILLM}                     & LLaVA-7B          & 2.57 & 2.64 & 3.13 & 2.21 & 2.39 \\
\textbf{RED-VILLM}                     & Qwen-VL-7B (en)   & 2.69 & 2.72 & 3.32 & 2.32 & 2.47 \\
\textbf{RED-VILLM}                     & Qwen-VL-7B (cn)   & 2.71 & 2.75 & 3.34 & 2.34 & 2.45 \\
\textbf{RED-VILLM}                     & Deepseek-VL-7B (cn) & \textbf{3.05} & \textbf{2.92} & \textbf{3.62} & \textbf{2.41} & \textbf{2.57} \\
\bottomrule
\end{tabular}
\end{table*}

\begin{table*}[h]
\centering
\caption{Comparison with leading methods on different zero-shot video QA datasets.}
\label{table: zero-shot qa}
\begin{tabular}{llcccccccc}
\toprule
\multirow{2}{*}{\textbf{Methods}} & \multirow{2}{*}{\textbf{LLM}} 
& \multicolumn{2}{c}{\textbf{MSVD-QA}} 
& \multicolumn{2}{c}{\textbf{MSRVTT-QA}} 
& \multicolumn{2}{c}{\textbf{TGIF-QA}} 
& \multicolumn{2}{c}{\textbf{ActivityNet-QA}} \\
\cmidrule(lr){3-4} \cmidrule(lr){5-6} \cmidrule(lr){7-8} \cmidrule(lr){9-10}
& & Accuracy & Score & Accuracy & Score & Accuracy & Score & Accuracy & Score \\
\midrule
FrozenBiLM~\cite{yang2022zero}       & DeBERTa-V2     & 32.2  & –   & 16.8  & –   & 41.0  & –   & 24.7  & –   \\
Video-LLaMA~\cite{zhang2023video}    & Vicuna-7B      & 51.6  & 2.5 & 29.6  & 1.8 & –     & –   & 12.4  & 1.1 \\
LLaMA-Adapter~\cite{zhang2023llama}  & LLaMA-7B       & 54.9  & 3.1 & 43.8  & 2.7 & –     & –   & 34.2  & 2.7 \\
VideoChat~\cite{li2023videochat}     & Vicuna-7B      & 56.3  & 2.8 & 45.0  & 2.5 & 34.4  & 2.3 & 26.5  & 2.2 \\
Video-ChatGPT~\cite{maaz2023video}   & LLaVA-7B       & 64.9  & 3.3 & 49.3  & 2.8 & 51.4  & 3.0 & 35.2  & 2.7 \\
\midrule
\textbf{RED-VILLM}                   & LLaVA-7B       & 68.9  & 2.8 & 52.4  & 2.9 & 55.9  & 3.1 & 39.2  & 3.0 \\
\textbf{RED-VILLM}                   & Qwen-VL-7B     & 71.2  & 3.7 & 53.9  & 3.1 & 62.3  & \textbf{3.3} & 44.2  & 3.2 \\
\textbf{RED-VILLM}                   & Deepseek-VL-7B & \textbf{73.5} & \textbf{3.8} & \textbf{55.7} & \textbf{3.8} & \textbf{64.1} & \textbf{3.3} & \textbf{48.6} & \textbf{3.5} \\
\bottomrule
\end{tabular}
\end{table*}

\subsection{Quantitative evaluation}
\textbf{Video-based Text Generation Performance.}  
We evaluate the generative performance of RED-VILLM, based on LLaVA, Qwen-VL and Deepseek-VL, using a video-based benchmark annotated from ActivityNet. This benchmark assesses the model on five aspects: Correctness of Information (CI), Detail Orientation (DO), Contextual Understanding (CU), Temporal Understanding (TU), and Consistency (CO). For Qwen-VL, we conduct tests in both Chinese and English, translating annotations and predictions as needed. The evaluation is performed using the GPT-3.5 pipeline, which assigns a score from 1-5 for each aspect.

As shown in Table~\ref{table: generative benchmark}, our RED-VILLM outperforms Video-ChatGPT across LLaVA, Qwen-VL and Deepseek-VL. Notably, the Chinese and English versions of Qwen-VL exhibit nearly identical performance, indicating that the language used has minimal impact on the GPT-3.5 evaluation.

\textbf{Zero-Shot Question-Answer Evaluation.}  
We evaluate the video question-answering capabilities of our model on four open-ended public datasets: MSRVTT-QA~\cite{xu2017video}, MSVD-QA~\cite{xu2017video}, TGIF-FrameQA~\cite{jang2017tgif}, and ActivityNet-QA~\cite{yu2019activitynet}. Evaluations are conducted in a zero-shot manner, with GPT-assisted scoring of answer relevance on a scale of 1-5.

Table~\ref{table: zero-shot qa} shows that RED-VILLM consistently outperforms Video-ChatGPT, demonstrating its superior video understanding and ability to generate accurate responses.

\begin{table}[htp]
\centering
\caption{Effect of temporal and spatial pooling for RED-VILLM.}
\label{table: effect sp features}
\resizebox{\linewidth}{!}{
\begin{tabular}{lccccccc}
\toprule
\textbf{Image LLM} & \textbf{T.P.} & \textbf{S.P.} & \textbf{CI} & \textbf{DO} & \textbf{CU} & \textbf{TU} & \textbf{CO} \\
\midrule
LLaVA-7B   & \ding{55} & \Checkmark & 2.12 & 2.20 & 2.37 & 1.64 & 2.08 \\
LLaVA-7B   & \Checkmark & \ding{55} & 2.28 & 2.35 & 2.54 & 1.79 & 2.21 \\
LLaVA-7B   & \Checkmark & \Checkmark & \textbf{2.40} & \textbf{2.52} & \textbf{2.62} & \textbf{1.98} & \textbf{2.37} \\
\midrule
Qwen-VL-7B & \ding{55} & \Checkmark & 2.27 & 2.43 & 2.82 & 1.65 & 2.12 \\
Qwen-VL-7B & \Checkmark & \ding{55} & 2.44 & 2.61 & 2.97 & 1.88 & 2.33 \\
Qwen-VL-7B & \Checkmark & \Checkmark & \textbf{2.51} & \textbf{2.69} & \textbf{3.01} & \textbf{2.11} & \textbf{2.43} \\
\bottomrule
\end{tabular}
}
\end{table}

\subsection{Ablation analysis}

\begin{figure*}
\centering
\includegraphics[width=1.0\textwidth]{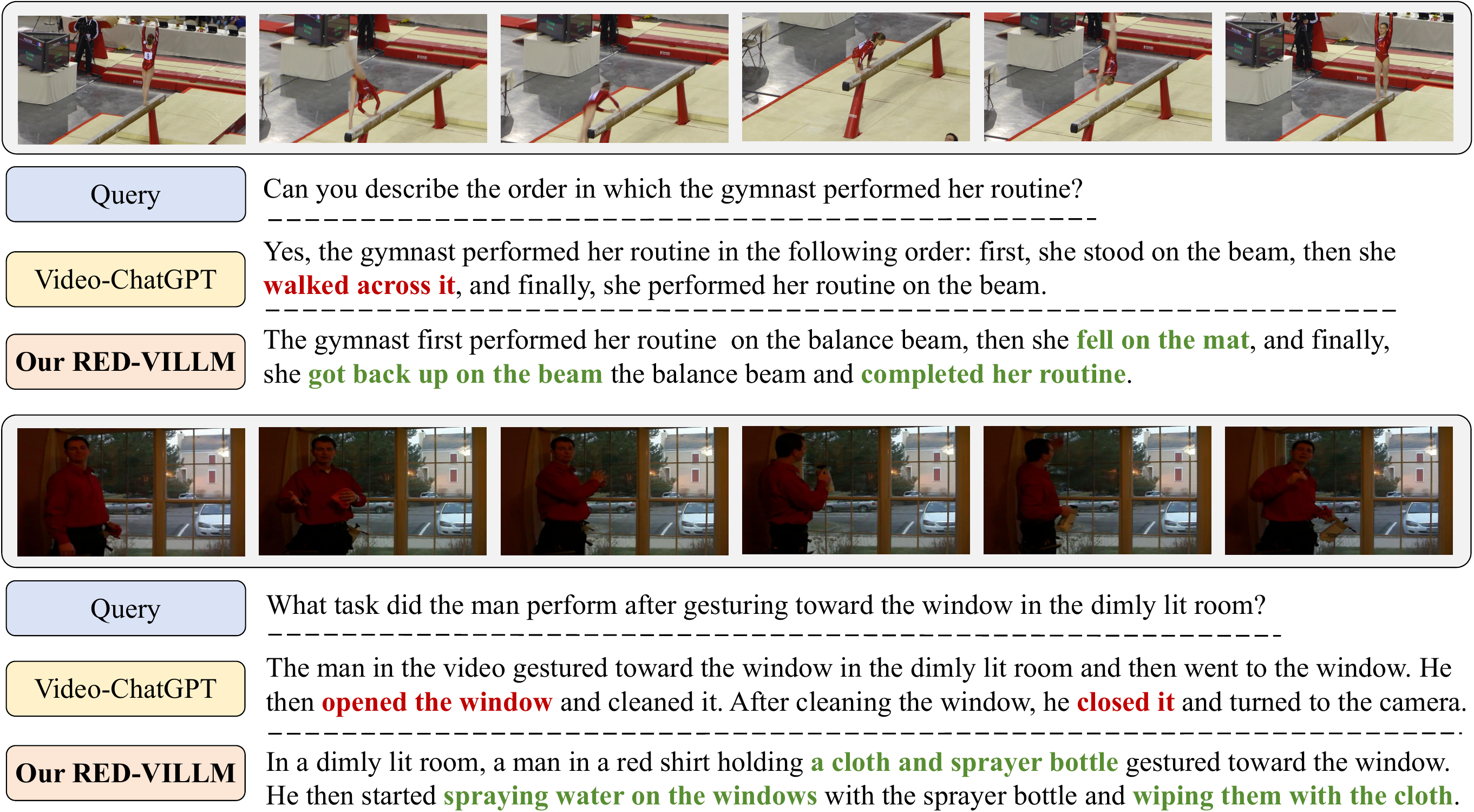}
\caption{Qualitative Results Comparison. We focus on comparing the differences in temporal understanding capabilities between RED-VILLM and the baseline, Video-ChatGPT.
{\color[RGB]{112,173,71}Green} signifies accurate descriptions, while {\color[RGB]{255,0,0}red} denotes incorrect or hallucinatory responses. It is evident that RED-VILLM significantly outperforms Video-ChatGPT in understanding temporal content.}
\label{fig:case}
\end{figure*}

\textbf{Effect of Temporal and Spatial Pooling.}
We investigate the impact of spatial and temporal pooling on video understanding (noted as S.P. and T.P. respectively). To isolate the effect of pooling, we retain the base Image LLM structure without the temporal module, using decoupled spatial and temporal features for training. We compare this setup with the coupled spatio-temporal features.

As shown in Table~\ref{table: effect sp features}, coupled spatio-temporal features yield the best performance across all five video understanding metrics. Using only spatial features (via temporal pooling) captures spatial details but lacks temporal understanding. Conversely, using only temporal features (via spatial pooling) reduces performance across all metrics due to the loss of spatial information. This highlights the importance of combining both spatial and temporal features for improved video understanding.

\begin{table}[htp]
\centering
\caption{Effect of plug-and-play temporal module for RED-VILLM.}
\label{table: temporal module}
\resizebox{\linewidth}{!}{
\begin{tabular}{lcccccc}
\Xhline{1pt}
\textbf{Image LLM}  & \makecell{\textbf{Temporal} \\ \textbf{Module}} & \textbf{CI} & \textbf{DO} & \textbf{CU} & \textbf{TU} & \textbf{CO} \\ \hline
LLaVA-7B   &    \ding{55}            &             2.40        & 2.52   & 2.62    & 1.98     & 2.37             \\
LLaVA-7B   &    \Checkmark             &            \textbf{2.57} & \textbf{2.64} &  \textbf{3.13}& \textbf{2.21}& \textbf{2.41}            \\ \hline
Qwen-VL-7B &     \ding{55}           &         2.51    &   2.69     &    3.01     &      2.11    &     2.43        \\
Qwen-VL-7B &      \Checkmark           &             \textbf{2.71}     &       \textbf{2.75} &      \textbf{3.34}   &    \textbf{2.34}      &       \textbf{2.45}         \\ \Xhline{1pt}
\end{tabular}
}
\end{table}

\textbf{Effect of Temporal Module.}
We further assess the impact of the plug-and-play temporal module by comparing it with the base Image LLM structure. In the base structure, spatio-temporal video features are directly aligned by the Image LLM's alignment module. In contrast, our method decouples spatio-temporal features and uses both the alignment module and the temporal module for better feature alignment.

As shown in Table~\ref{table: temporal module}, the temporal module significantly improves video understanding, particularly temporal comprehension. This confirms that the temporal module enhances the Image LLM's ability to process temporal information, resulting in a more comprehensive understanding of video content. Notably, when the base model is LLaVA, RED-VILLM with the temporal module outperforms Video-ChatGPT, demonstrating the effectiveness of the temporal module.

\begin{table}[tbp]
\centering
\caption{Effect of additional MLP for RED-VILLM.}
\label{table: additional MLP}
\adjustbox{max width=\linewidth}{
    \begin{tabular}{lccccc}
    \toprule
    \textbf{Add. MLP} & \textbf{CI} & \textbf{DO} & \textbf{CU} & \textbf{TU} & \textbf{CO} \\
    \midrule
    with   & 2.35 & 2.41 & 2.62 & 1.59 & 2.01 \\
    w/o    & \textbf{2.71} & \textbf{2.75} & \textbf{3.34} & \textbf{2.34} & \textbf{2.45} \\
    \bottomrule
    \end{tabular}
}
\end{table}

\textbf{Design of Temporal Module}
The temporal module in our pipeline is initialized with the pre-trained parameters of the base Image LLM alignment module to maintain its effectiveness. Adding randomly initialized parameters could negatively impact the alignment process.

We verify this by training Video LLMs based on Qwen-VL with and without an additional MLP layer behind the alignment module. As shown in Table~\ref{table: additional MLP}, adding the MLP layer significantly degrades performance across all metrics, confirming that introducing random parameters disrupts the alignment process. Thus, our method benefits from keeping the temporal module structure aligned with the pre-trained parameters of the base Image LLM alignment module.

\subsection{Qualitative Analysis}
As shown in Fig.~\ref{fig:case}, we compare the performance of Video-ChatGPT and RED-VILLM, both based on LLaVA. Video-ChatGPT sometimes fails to correctly order events, leading to incoherent or inaccurate descriptions. In contrast, RED-VILLM consistently captures and accurately represents these temporal relationships. This highlights the critical role of the temporal module in improving RED-VILLM's temporal understanding, especially for videos with significant temporal changes.

\section{Conclusion}
The conventional two-stage training paradigm of Video LLM (i.e., pre-training and fine-tuning) suffers from expensive training data and neglects the potential of ready-made Image LLM. Therefore, in this paper, we propose RED-VILLM, the Resource-Efficient Development pipeline to build Video LLM by leveraging the prior knowledge on text-image alignment embedded in pre-trained Image LLM. Through introducing a plug-and-play temporal adaptation module, we efficiently extend Image LLMs' capabilities to dynamic video content. Our approach realizes a robust Video LLM while significantly reducing training data and computing resources requirements.

Extensive experiments demonstrate the superiority of our proposed pipeline. Our study makes the development of Video LLMs more accessible and opens pathways for further researches in resource-efficient methods in multimodal learning.

{
    \small
    \bibliographystyle{ieeenat_fullname}
    \bibliography{main}
}

\appendix
\begin{figure*}[t]
\includegraphics[width=1.0\textwidth]{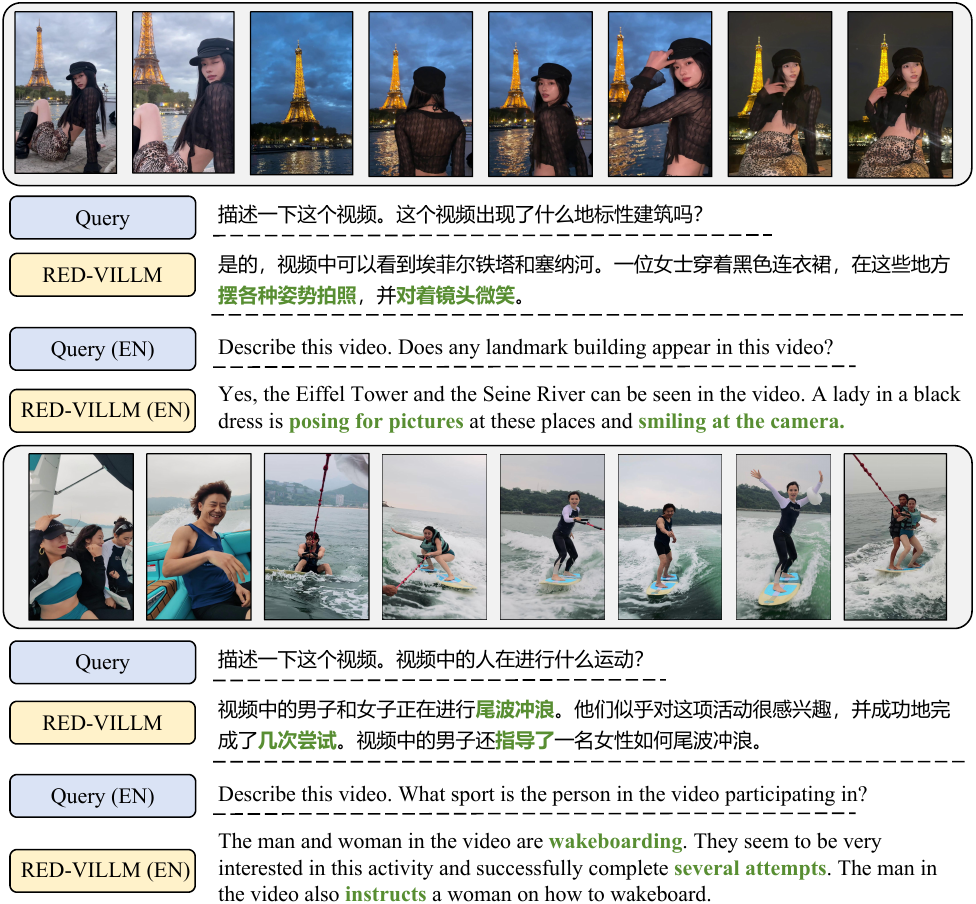}
\caption{Qualitative Results in Chinese. We present the Chinese capabilities of our RED-VILLM in video understanding. "EN" represents the corresponding English translation of the Chinese content.}
\label{fig:cncase}
\end{figure*}

\section{Technical Details}
\subsection{Resource-Efficient Nature}
In our approach, we emphasize the efficiency and resource-saving nature. Since our proposed RED-VILLM is based on a pre-trained Image LLM, we do not require extensive multimodal alignment data nor the need to finetune the LLM. Instead, we only need to train the alignment module of the base Image LLM with few video instruction data.

\begin{table}[h]
\centering
\caption{Comparison with leading methods on training data and trainable parameters.}
\label{table: efficient}
\resizebox{\linewidth}{!}{
\begin{tabular}{lcc}
\toprule
\textbf{Methods} & \textbf{Training Data} & \textbf{Trainable Parameters} \\
\midrule
Video-LLaVA        & 2025K multimodal data        & 7B \\
LLaMA-VID          & 1568K multimodal data        & 7B \\
RED-VILLM (LLaVA)   & 100K video instructions data & 0.13B \\
RED-VILLM (Qwen-VL) & 100K video instructions data & 0.7B \\
\bottomrule
\end{tabular}
}
\end{table}

As shown in Table~\ref{table: efficient}, compared to Video-LLaVA, our RED-VILLM based on LLaVA or Qwen-VL leads to fewer trainable parameters (2\% and 10\% respectively) and utilizes only 5\% of the training data.

\begin{table}[h]
\centering
\caption{Comparison with leading methods on different zero-shot video QA datasets with significantly reduced training data.}
\label{table: baseline}
\resizebox{\columnwidth}{!}{
\begin{tabular}{llcccccccc}
\toprule
\multirow{2}{*}{\textbf{Methods}} & \multirow{2}{*}{\textbf{LLM}} 
& \multicolumn{2}{c}{\textbf{MSVD-QA}} 
& \multicolumn{2}{c}{\textbf{MSRVTT-QA}} 
& \multicolumn{2}{c}{\textbf{TGIF-QA}} 
& \multicolumn{2}{c}{\textbf{ActivityNet-QA}} \\
\cmidrule(lr){3-4} \cmidrule(lr){5-6} \cmidrule(lr){7-8} \cmidrule(lr){9-10}
& & Accuracy & Score & Accuracy & Score & Accuracy & Score & Accuracy & Score \\
\midrule
Video-LLaVA   & Vicuna-7B    & 70.7 & \textbf{3.9} & \textbf{59.2} & \textbf{3.5} & \textbf{70.0} & \textbf{4.0} & 45.3 & \textbf{3.3} \\
LLaMA-VID     & Vicuna-7B    & 69.7 & 3.7          & 57.7          & 3.2          & –             & –           & \textbf{47.4} & \textbf{3.3} \\
\midrule
RED-VILLM     & LLaVA-7B     & 68.9 & 2.8          & 52.4          & 2.9          & 55.9          & 3.1         & 39.2 & 3.0 \\
RED-VILLM     & Qwen-VL-7B   & \textbf{71.2} & 3.7    & 53.9          & 3.1          & 62.3          & 3.3         & 44.2 & 3.2 \\
\bottomrule
\end{tabular}
}
\end{table}

However, as shown in Table~\ref{table: baseline}, under significantly reduced training costs and training data, our RED-VILLM achieves comparable performance to Video-LLaVA and LLaMA-VID, which demonstrates the efficiency of our approach.

\subsection{Training Procedure}
For input video, our approach employs a uniform sampling strategy, making it agnostic to the FPS of the input video. Specifically, we sample 100 frames from the input. Subsequently, we leverage Vision Transformer (ViT) with a resolution of 224×224 for video encoding. Eventually, spatial and temporal pooling are applied to project the visual representations into 356 tokens (256 and 100 respectively).

We train RED-VILLM on the 100k dataset over 3 epochs. For such training, RED-VILLM based on LLaVA or Qwen-VL requires approximately 3 and 4.5 A100 hours correspondingly. During inference, we employ FP16 mode. Taking the ActivityNet QA test set with 8k videos as an example, inference with RED-VILLM based on LLaVA or Qwen-VL requires 1.2 and 2 hours respectively, averaging 0.54 and 0.9 seconds per item.

We also performed experiments on sampling in Table~\ref{table: num frames}, to validate our sampling strategy.

\begin{table}[h]
\centering
\caption{Effect of the number of sampled frames on performance.}
\label{table: num frames}
\begin{tabular}{lccccc}
\toprule
\textbf{Frames} & \textbf{CI} & \textbf{DO} & \textbf{CU} & \textbf{TU} & \textbf{CO} \\
\midrule
50   & 2.39 & 2.52 & 2.92 & 2.09 & 2.42 \\
75   & 2.53 & 2.67 & 3.11 & 2.21 & 2.40 \\
100  & \textbf{2.71} & \textbf{2.75} & \textbf{3.34} & \textbf{2.34} & 2.45 \\
150  & 2.65 & 2.65 & 3.26 & 2.29 & \textbf{2.47} \\
200  & 2.57 & 2.64 & 3.24 & 2.21 & 2.44 \\
\bottomrule
\end{tabular}
\end{table}

\subsection{Design of Temporal Module}
To endow the Image LLM with the capability to grasp temporal information, we design the Temporal Module. Note that we emphasize the efficiency and resource-saving property of developing a robust Video LLM. We delve with various Temporal Module designs to find the most efficient structure.

After extensive experiments, we discover that initializing the Temporal Module with the structure and parameters of the base Image LLM alignment module results in the most efficient training and the most competitive performance in understanding temporal information. We infer that the Image LLM already accomplishes substantial alignment work. Therefore, starting from the ordinary alignment module is more efficient than starting from scratch.

\section{Qualitative Results in Chinese}
In the main text, we already showcase the video comprehension abilities of our proposed RED-VILLM in English. In this section, we select some videos from the Chinese community and present the performance of our RED-VILLM using Qwen-VL as backbone Image LLM. Note that we query RED-VILLM in Chinese and translate the Chinese questions and responses into English for presentation.

As shown in Fig.~\ref{fig:cncase}, our approach demonstrates excellent video understanding capabilities, particularly in terms of its proficiency with Chinese content. Based on RED-VILLM, we achieve the first Video LLM designed for understanding videos within the Chinese-speaking community, requiring minimal instructional data and training resources. Our proposed pipeline makes development of Video LLMs more accessible for both academia and industry.
\end{document}